\documentclass[a4paper,11pt,article,oneside]{memoir}
\usepackage{ecis2024}
\usepackage{enumitem}
\usepackage{amsmath}
\usepackage{svg}
\usepackage{amsthm}
\usepackage[capitalize,noabbrev]{cleveref}
\newtheoremstyle{exampstyle}
  {\topsep} 
  {\topsep} 
  {} 
  {} 
  {\bfseries} 
  {.} 
  {.5em} 
  {} 
\theoremstyle{exampstyle}\newtheorem{hyp}{Hypothesis}

\newlist{questions}{enumerate}{2}
\setlist[questions,1]{label=RQ:}

\newlist{worddefs}{description}{1}

\maintitle{\textsc{CollaFuse}: Navigating Limited Resources and Privacy in Collaborative Generative AI} 
\shorttitle{Collaborative Diffusion Models} 
\category{Short Paper} 

\authors{
Domenique Zipperling, University of Bayreuth and Fraunhofer FIT, Bayreuth, Germany, \\
\hangindent=0.5cm domenique.zipperling@uni-bayreuth.de

Simeon Allmendinger, University of Bayreuth and Fraunhofer FIT, Bayreuth, Germany, \\ 
\hangindent=0.5cm simeon.allmendinger@uni-bayreuth.de

Lukas Struppek, German Center for Artificial Intelligence \& TU Darmstadt, Darmstadt, Germany, \\ 
\hangindent=0.5cm lukas.struppek@cs.tu-darmstadt.de

Niklas Kühl, University of Bayreuth and Fraunhofer FIT, Bayreuth, Germany, \\ 
\hangindent=0.5cm kuehl@uni-bayreuth.de
}

\abstracttext{In the landscape of generative artificial intelligence, diffusion models present challenges for socio-technical systems in data requirements and privacy. Traditional approaches like federated learning distribute the learning process but strain individual clients, especially with constrained resources. In response to these challenges, we introduce \textsc{CollaFuse}, a novel framework inspired by split learning. Tailored for efficient and collaborative use of denoising diffusion probabilistic models, \textsc{CollaFuse} enables shared server training and inference, alleviating client computational burdens. This is achieved by retaining data and computationally inexpensive GPU processes locally at each client while outsourcing the computationally expensive processes to the shared server. Demonstrated in a healthcare context, \textsc{CollaFuse} enhances privacy by reducing the need for sensitive information sharing. These capabilities hold the potential to impact various application areas, such as edge computing, healthcare research, or autonomous driving. In essence, our work advances distributed machine learning, shaping the future of collaborative GenAI networks.
}


\begin{document}

\chapter{Introduction}
\label{introduction}

In the realm of Information Systems Research, the emergence of generative artificial intelligence (GenAI) technologies like ChatGPT and DALL-E has marked a significant milestone. These advancements, as detailed by~\citet{feuerriegel2023generative}, have broadened public access to GenAI and catalyzed its integration into diverse sectors. Among GenAI innovations, the denoising diffusion probabilistic model (DDPM) stands out for its ability to generate high-quality images through an advanced denoising process, outperforming earlier methods such as generative adversarial networks (GANs)~\citep{goodfellow20gans} or variational autoencoders (VAEs)~\citep{Kingma2014VAE} in terms of diversity and convergence guarantees~\citep{nichol21improving, dhariwal21beating}.
However, the implementation of DDPMs in business analytics and other fields is not without challenges. These models demand extensive data sets and computational resources~\citep{wang2023patch}, which are often limited, especially in decentralized systems~\citep{hirt2019cognition}. The healthcare sector exemplifies these constraints, where data scarcity, privacy concerns, and high costs of data collection are present~\citep{Veeraragavan2023SecuringFLGAN, Wang2023PrivacyFL_ICU}. Nonetheless, DDPMs possess significant potential in the healthcare sector for example in medical image synthesis, reconstruction, or Image-to-Image translation.~\citep{KAZEROUNI2023102846}. To address these challenges, researchers have explored various strategies, including patch-wise training~\citep{wang2023patch}, few-shot learning ~\citep{Lu2023specialist, ruiz2023dreambooth, Zhang2023Sine}, and notably, federated learning (FL)~\citep{McMahan2017FL, fan2020federated}. FL enhances data accessibility, yet raises privacy concerns~\citep{shokri17meminf,zhu19leakage} and the need for significant local computational capabilities.

In response, we introduce \textsc{\textbf{CollaFuse}}, a new collaborative learning and inference framework for DDPMs, inspired by split learning (SL)~\citep{gupta2018distributed}. \textsc{CollaFuse} aims to balance the computationally intensive denoising process between local clients and a shared server, with a focus on optimizing the trade-off between performance, privacy, and resource utilization---which are crucial requirements for real-world information systems implementations. This framework transforms the optimization challenge into a multi-criteria problem, addressing the core requirements of such applications in practice. Building upon these criteria, our research investigates the impact of different degrees of collaboration defined by a cut-ratio $c \in [0,1]$, formulating two key hypotheses:

\begin{hyp}
    In our framework \textsc{CollaFuse}, collaborative learning of DDPMs positively influences the fidelity of generated images compared to non-collaborative local training ($c=1$).
\label{hyp1}
\end{hyp}
\begin{hyp}
    Increasing collaborative effort ($c\downarrow$) improves performance (a) and the amount of disclosed information (b) while implicitly reducing the locally consumed GPU energy (c).
\label{hyp2}
\end{hyp} 

Our initial analysis, including a healthcare-focused experiment with magnetic resonance imaging (MRI) brain scans, supports these hypotheses. As a consequence, \textsc{CollaFuse} holds promise for applicants such as small medical institutions or even individual practitioners with edge devices to engage in collaborative model training and inference, e.g., medical training. Beyond healthcare, the framework exhibits potential in domains like autonomous driving, where edge computing resources are constrained, yet computational demands and privacy considerations are high. 
Moving forward, our research will delve deeper into the \textsc{CollaFuse} framework, analyzing its performance in terms of image fidelity and diversity, assessing potential privacy risks, and exploring resource efficiency in additional scenarios. 
This comprehensive investigation aims to advance our understanding of (distributive) GenAI within socio-technical systems as demanded by related literature~\citep{feuerriegel2023generative,abbasi2023data}. Consequently, our ongoing research on \textsc{CollaFuse} can offer guidelines on how to apply DDPMs collaboratively and work as a blueprint for future GenAI networks designs in various domains.

\chapter{Background and Related Work}
\label{background}
Our research is grounded in the collaborative concepts of distributed learning and the architectural principles of diffusion models. \textit{FL} was first introduced by \citet{McMahan2017FL} utilizing distributed data for training without storing it centrally. Since then, FL has been on a triumphant march and received a lot of attention~\citep{hard2019federated}. At its core, FL iteratively composes locally trained models into a global model requiring clients to share model updates and gradients, which increases the risk of data leakage~\citep{shokri17meminf,zhu19leakage}. Furthermore, FL comes with high computational requirements at the client-side~\citep{SplitFed2022}.
Another paradigm for collaborative learning is \textit{SL} exploiting the sequencing of operations in neural networks~\citep{gupta2018distributed}. In general, SL splits a neural network among multiple clients and a shared server. This is especially intriguing as clients can use a server to train most of the model while keeping the data and labels locally. 
In 2020, research advanced diffusion models with the introduction of \textit{Denoising Diffusion Probabilistic Models} (DDPM)~\citep{Ho2020}, offering an alternative to GANs~\citep{goodfellow20gans} for image generation. DDPMs involve two processes: diffusion and denoising. The diffusion process adds noise from a Gaussian distribution incrementally to an image over $T$ steps. In the denoising process, the model estimates the noise added at each step $t \in [0,T]$. While training the model, weights are updated, calculating the loss based on the difference between true and estimated noise. Accordingly, the initial training image is not necessary for the denoising process. On this basis, DDPMs are able to generate new images from pure Gaussian noise, which closely resemble the images of the training data set. Fidelity metrics, like the Kernel-Inception distance (KID)~\citep{binkowsi18kid}, gauge diffusion model performance by quantifying the difference between the distributions of real and generated images.


Recent developments in GenAI have spurred interest within the information systems (IS) community. ~\citet{feuerriegel2023generative} and~\citet{Banh2023GenAI} broadly define GenAI, covering its architectures and applications, while identifying challenges and research questions for the IS community. Additionally, research has addressed how GenAI, particularly the language generation model ChatGPT, influences education marketing or intellectual property~\citep{Peres2023, Burger2023}.
Moreover, FL has also captured the attention of the IS community.~\citet{karnebogen2023SLR_FL} have introduced a taxonomy for FL applications, while~\citet{Wang2023PrivacyFL_ICU} have investigated FL for privacy-preserving healthcare applications.~\citet{hirt2023enabling} explore the impacts of distributed learning on organizations. However, the integration of FL with GenAI, particularly in image synthesis, remains unexplored within IS research, prompting a look towards approaches from other communities, such as computer science. Within the computer science community GANs have been used for a long time to synthesize images. As GANs are composed of two components, a generator and discriminator, research has been centered around whether both or only one component should be trained collaboratively. \citet{Hardy2019MD-GAN, wang2023fedmedgan} propose only to train the generator collaboratively and to train the discriminator at each client. In contrast, \citet{fan2020federated} offer empirical results indicating that synchronizing discriminator and generator across clients yields the best results, leading to the proposal of an improved FL-Gan version~\citep{Li2022IFL-GAN}.
As FL exhibits privacy concerns, \citet{augenstein2020generative} apply differential privacy while \citet{Veeraragavan2023SecuringFLGAN} add consortium blockchains and Shamir's Secret Sharing algorithm to federated GANs. Additionally, \citet{Benshun2023SL-GAN} propose a federated split learning framework showing efforts to combine FL and SL.
Research on collaborative training methods is still scarce in the domain of \textit{Diffusion Models}. \citet{jothiraj2023phoenix} introduce Phoenix, a method for training diffusion models in federated learning to address data diversity issues, sharing minimal data (4-5\%) among clients to cut communication costs. It excels in precision and recall but needs further development to enhance image quality.
Current research on collaborative GenAI largely overlooks the advantages of other DDPMs and SL, which can offer reduced resource use and enhanced privacy. With \textsc{CollaFuse}, we aim to leverage these benefits to advance GenAI.

\chapter{Framework}
\label{framework}
\begin{figure}[t]
    \centering
    \includegraphics[width=\textwidth]{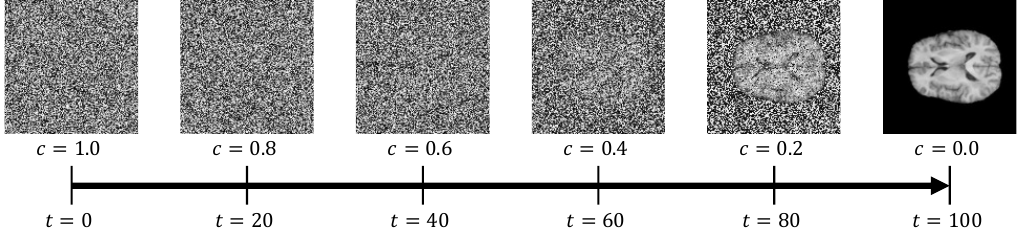}
    \caption{Illustration of the denoising process in DDPMs: Exemplary images generated at denoising step $t$ for various cut-ratios $c$. The distinguishing features of the generated images remain effectively concealed behind noise during the majority of denoising steps.}\label{fig:denoising_process}  
\end{figure}

We propose \textsc{CollaFuse}, a framework facilitating collaborative GenAI for image generation with DDPMs across clients. Inspired by SL, the less resource-intensive diffusion process is computed locally by each client, whereas the computationally intensive denoising process is strategically split at step $t_c = (1-c)T$ during both training and inference. This results in a shared model (backbone) hosted on a shared server, coupled with individual local models for each client. The unique design of \textsc{CollaFuse} allows clients to retain sensitive data locally while outsourcing the majority of computationally intensive denoising operations to a centralized server. The split is governed by the cut-ratio $c$, dictating the computational load for each client and the extent of disclosed information. \Cref{fig:denoising_process} shows this governance, where, for instance, with $T=100$ and $c=0.8$, the first 20\% of the denoising process (from $t=1$ to $t_{0.8}=20$) is managed on a shared server and the subsequent 80 steps (80\%) are locally executed, maintaining the privacy of these stages
As illustrated in \Cref{fig:process_collaborative_diffusion_models}a, from the client's perspective, the training sequence orchestrated by \textsc{CollaFuse} comprises six key steps. The server initiates the client's diffusion process (1), after which the client computes its image data's diffusion process batch-wise (2) and forwards the resulting noised images, along with the corresponding added random noise $\epsilon=(\epsilon_{0},..., \epsilon_{t_c})$ for each image, back to the server (3). Utilizing the noised images, the server undertakes the initial phase of the denoising process, computing the loss between the estimated and provided random noise (4). The partially denoised images are subsequently transmitted back to the client (5), where the remaining denoising process is executed (6). 
The system perspective is depicted in \Cref{fig:process_collaborative_diffusion_models}b, showcasing two clients providing noised images to the shared server and receiving partially denoised images for further training. Further, clients can generate new images by sampling images from pure random noise and passing them to the server.

\chapter{Experimental Evaluation}
\label{method}
To assess our framework, \textsc{CollaFuse}, we simulate a healthcare-related scenario involving three clients and one server. Every client data set is independent comprising 4,920 MRI scans from 123 patients each. The hold-out test data set contains 5,000 images from 125 further patients~\citep{Bakas2017}. Our diffusion model employs a U-Net architecture~\citep{ronneberger15unet} including ResNet~\citep{resnet_he} blocks for down- and upsampling and self-attention mechanisms~\citep{vaswani17attention} for feature refinement and textual guidance. Furthermore, we employ an identical cosine variance scheduler, $T = 100$ steps, and maintain a fixed image size of ($128\times128$) across clients. The training process spans $300$ epochs with a fixed learning rate of $0.001$ and a batch size of $150$.
Our investigation delves into the impact of the cut-ratio $c$ on the trade-off between performance, disclosed information, and GPU energy consumption considering cut-ratio values $c \in [0.0, 0.2, ..., 1.0]$. Performance is assessed using the common fidelity metric KID~\citep{binkowsi18kid} on both the client-dependent training and hold-out data sets, employing the feature extractor from the clean-fid library~\citep{parmar_On}. GPU energy consumption is measured using the \textit{codecarbon} Python package~\footnote{https://mlco2.github.io/codecarbon/}. Disclosed information for each client is approximated through the mean squared error (MSE) for a pixel-by-pixel comparison and KID scores between partially denoised images at the split step $t_c$ and real images of clients.
To simulate a distributed healthcare-related scenario, we utilize a cluster of four NVIDIA A100-SXM4-40GB GPUs. The implementation of the architecture, framework and experiment can be found here: \url{https://github.com/SimeonAllmendinger/collafuse.git.}
\begin{figure}[t]
    \centering{\includegraphics[clip, scale=0.45]{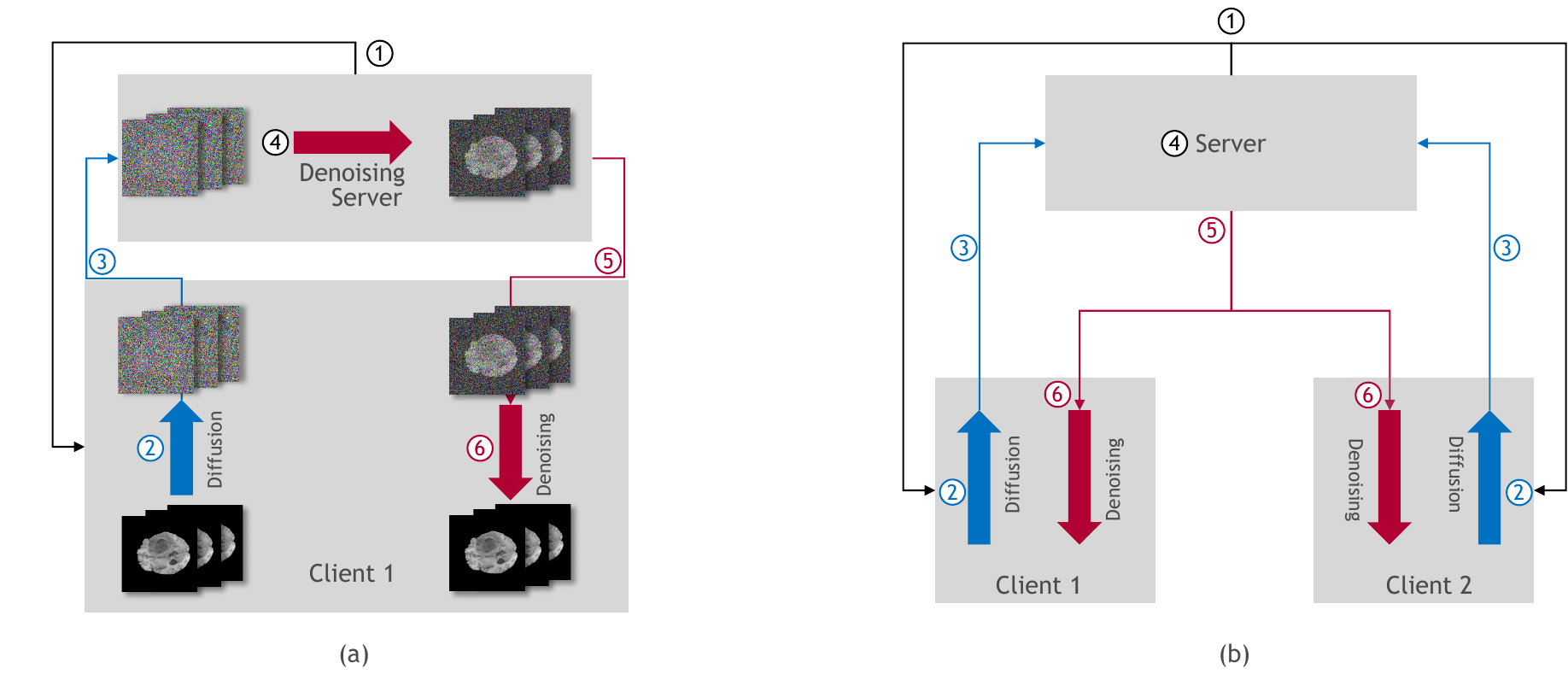}}
    \caption{The training procedure of \textsc{CollaFuse} comprises six steps from client perspective (a) and system (b): Server triggers diffusion process of clients (1), clients apply diffusion (2), clients send diffused images and noise to server (3), server denoises the images until $t_c$ (4), server sends the partially denoised images to client (5), and clients locally finish denoising process (6).}
    \label{fig:process_collaborative_diffusion_models} 
\end{figure}
\chapter{Results}
\label{results}
\begin{figure}[t]
    \centering
    \includegraphics[width=\textwidth]{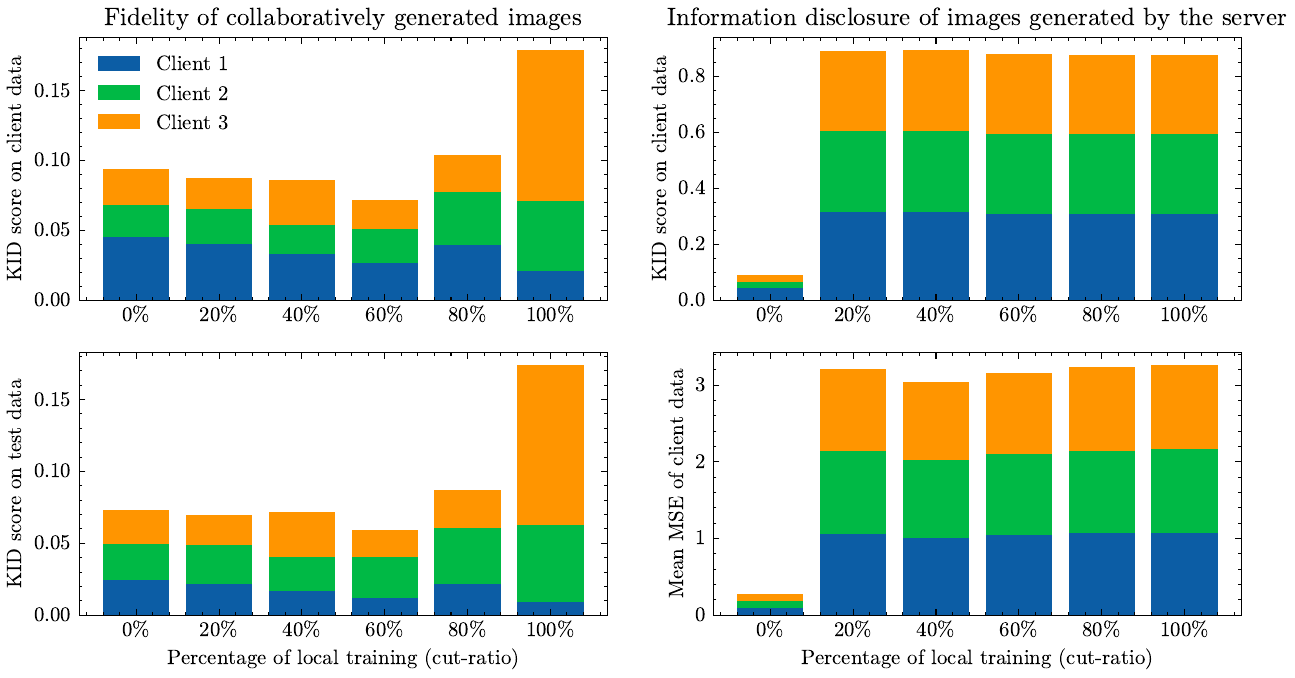}
    \caption{Preliminary results: We calculate
    the KID score to assess generative performance and out-of-sample robustness (upper-left, lower-left).
    Information disclosure is evaluated by comparing original client data to server-generated images, using
    KID for distribution (upper-right) and MSE for pixel-level comparison. Despite conducting up to 80\% of the computationally intensive denoising steps on the server, a significant amount of information in the images remains concealed.}
    \label{fig: Performance_and_disclosed_information}
\end{figure}

Our findings analyze trade-off among all three dimensions: the performance, gauged by the fidelity of generated images; the extent of disclosed information, verifying whether images at the split point equal original images within respective client data sets; and the GPU power usage across clients and server. \Cref{fig: Performance_and_disclosed_information} illustrates the trade-off between performance and disclosed information. We calculate the KID score to assess generative performance and out-of-sample robustness (upper-left, lower-left). Information disclosure is evaluated by comparing original client data to server-generated images, using KID for distribution analysis (upper-right) and MSE for pixel-level comparison. Lower KID scores signify enhanced performance, underscoring fidelity in generated images. Conversely, for disclosed information the objective is to maximize KID and MSE scores, ensuring the revelation of as few characteristics as possible on the shared server.
The analysis of \Cref{fig: Performance_and_disclosed_information} unveils two significant observations. Firstly, the stacked KID scores of performance exhibit a U-shaped pattern concerning the cut-ratio, particularly evident for client data. Consequently, collaborative efforts may lead to a reduced aggregated KID score compared to training the models locally ($100\%$). Interestingly, the performance is reduced, if a large part of the global denoising process is conducted on the shared server ($0\%-40\%$), thereby contributing to the observed U-shaped trend. With regard to the disclosed information, both pixel-wise comparison and the KID score imply that, despite conducting up to 80\% of computationally intensive denoising steps on the server, a substantial portion of information associated with the images remains concealed in comparison to total global denoising ($0\%$). Moreover, GPU power usage of the diffusion process exhibits limited computational intensity in the experiment, and the relocation of denoising steps to the server correlates positively with reduced local GPU energy demand. Overall, our experiment supports Hypothesis 1, showing that collaborative learning with \textsc{CollaFuse} improves image fidelity compared to non-collaborative local training (c = 1). Hypothesis \ref{hyp2} is evident in disclosed information, and reduced local computational intensity when denoising steps are moved to the server. Fascinatingly, concerning performance, the hypothesis only holds to some extent as the results indicate a tipping point where further collaboration does not lead to an increase in performance.

\chapter{Conclusion and Outlook}
\label{conclusion}
In this paper, we introduce \textsc{CollaFuse}, an innovative collaborative learning and inference framework designed for denoising diffusion probabilistic models. The primary objective is to address the trade-off between performance, privacy, and resource utilization---an imperative aspect for the practical implementation of information systems in real-world scenarios. Drawing inspiration from split learning~\citep{gupta2018distributed}, \textsc{CollaFuse} aims to balance the computationally intensive denoising process across local clients and a shared server. The framework is particularly beneficial in domains where data is scarce, private, and computational resources on local devices are limited. This especially includes scenarios of industry~\citep{li2022future} and healthcare~\citep{sivarajah2023responsible}. \textsc{CollaFuse} innovatively partitions the computationally extensive denoising process into two independently trainable components. The latter remains with the client, ensuring data privacy, while the initial part is trained collaboratively on a server, amplifying the amount of training data. We demonstrate that clients can execute numerous denoising steps on the server before data is disclosed. The findings further indicate that a decreasing cut-ratio $c$ effectively shifts computational effort to a shared server backbone, enhancing performance generalizability. In summary, our experiment provides initial evidence supporting the advantages of collaborative learning within \textsc{CollaFuse} across performance, disclosed information, and local GPU energy. \textsc{CollaFuse} offers a technical solution for handling sensitive and scarce data, creating new opportunities for the IS community by exploring collaborative GenAI across fields like healthcare, edge devices, and the automotive industry.
Looking ahead, our research roadmap involves a more in-depth analysis of \textsc{CollaFuse}. We will increase complexity by shifting to color imagery, expanding attribute diversity, and integrating text-driven generation with the Imagen model. Furthermore, we will address security by exploring attack vectors and implementing safeguards like differential privacy. This thorough examination will contribute to a deeper understanding of the transformative potential of \textsc{CollaFuse} influencing applications in industry and healthcare and paving the way for future advancements in collaborative generative AI applications. A promising field of research lies ahead.

\printbibliography

\end{document}